\documentclass[10pt,twocolumn,letterpaper]{article}

\usepackage[T1]{fontenc}
\usepackage[utf8]{inputenc}
\usepackage[pagenumbers]{wacv} % To force page numbers, e.g. for an arXiv version
\usepackage{booktabs}
\usepackage{multirow}
\usepackage{makecell}
\usepackage{graphicx}
\usepackage{adjustbox}

\definecolor{wacvblue}{rgb}{0.21,0.49,0.74}
\usepackage[pagebackref,breaklinks,colorlinks,allcolors=wacvblue]{hyperref}

\def\wacvPaperID{-} % *** Enter the Paper ID here
\def\confName{-}
\def\confYear{2026}

\title{SkinMap: Weighted Full-Body Skin Segmentation for Robust Remote Photoplethysmography}

\author{
    Zahra Maleki\thanks{Contributed equally to this work}\quad
    Amirhossein Akbari\footnotemark[1]\quad
    Amirhossein Binesh\quad
    Babak Khalaj\\
    {\tt\small \{zahra.maleki, amirhoseinakbari\}@ee.sharif.edu, ah.binesh@ce.sharif.edu, khalaj@sharif.edu}
 \\
    Sharif University of Technology
}

\begin{document}

\maketitle
\begin{abstract}

Remote photoplethysmography (rPPG) is an innovative method for monitoring heart rate and vital signs by using a simple camera to record a person, as long as any part of their skin is visible. This low-cost, contactless approach helps in remote patient monitoring, emotion analysis, smart vehicle utilization, and more. Over the years, various techniques have been proposed to improve the accuracy of this technology, especially given its sensitivity to lighting and movement. In the unsupervised pipeline, it is necessary to first select skin regions from the video to extract the rPPG signal from the skin color changes. We introduce a novel skin segmentation technique that prioritizes skin regions to enhance the quality of the extracted signal. It can detect areas of skin all over the body, making it more resistant to movement, while removing areas such as the mouth, eyes, and hair that may cause interference. Our model is evaluated on publicly available datasets, and we also present a new dataset, called SYNC-rPPG, to better represent real-world conditions. The results indicate that our model demonstrates a prior ability to capture heartbeats in challenging conditions, such as talking and head rotation, and maintain the mean absolute error (MAE) between predicted and actual heart rates, while other methods fail to do so. In addition, we demonstrate high accuracy in detecting a diverse range of skin tones, making this technique a promising option for real-world applications.
\end{abstract}
    
\section{Introduction}
\label{sec:intro}

Remote photoplethysmography (rPPG) is an advanced non-contact technique that enables the measurement of vital physiological signals \cite{a1}, such as heart rate (HR), respiratory frequency (RF), and heart rate variability (HRV), by analyzing video captured from any part of the skin surface. The light reaching the camera sensor has an AC component that reflects variations in light absorption caused by changes in arterial blood volume \cite{a2,a3,a4}. Unlike traditional contact-based sensors such as PPG or ECG, which require specialized equipment that can be costly and inaccessible \cite{a5}, rPPG offers a scalable and non-invasive solution for user monitoring. This technology holds significant promise for applications in remote healthcare, emotion analysis, and facial security \cite{a6}, as it can capture data from any exposed area of the skin without requiring physical proximity.

\begin{figure*}
  \centering
   \includegraphics[width=\linewidth]{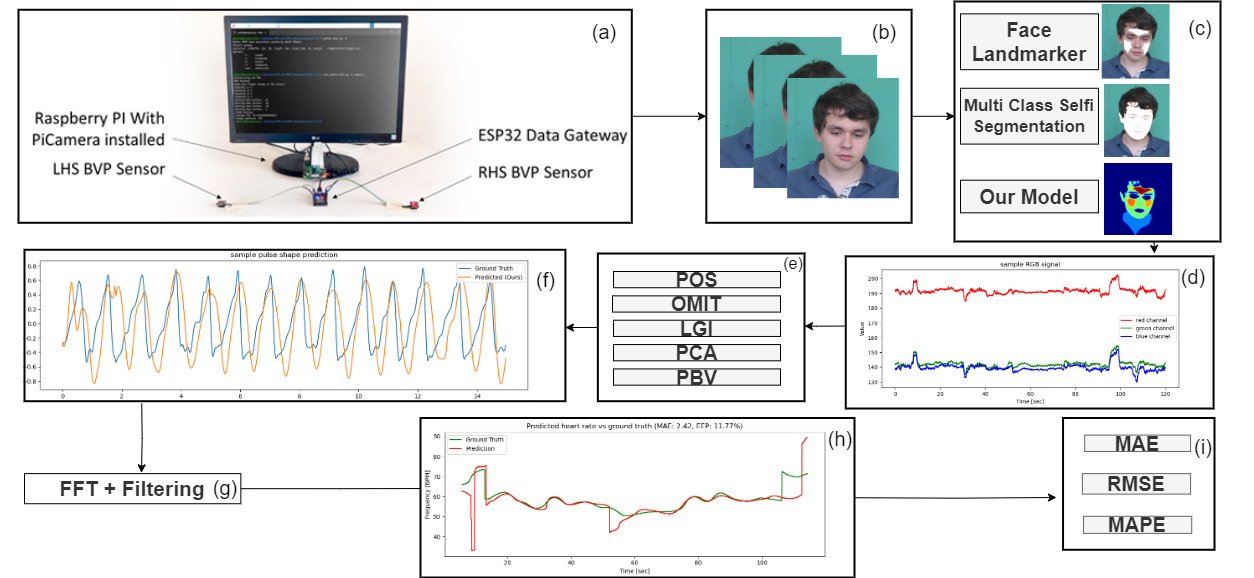}
\caption{Unsupervised pipeline for heart rate estimation from video. 
(a) Data acquisition.
(b) Video dataset collection synchronized with PPG signals.  
(c) Skin segmentation or ROIs selection process.
(d) RGB signal extraction by averaging skin pixels.  
(e) rPPG signal extraction methods are applied to the RGB signal.  
(f) Comparison of the extracted rPPG signal with the reference PPG pulse.  
(g) Heart rate estimation.  
(h) Heart rate analysis over time.  
(i) Evaluation of our estimation using statistical metrics.  }
   \label{fig:pipeline}
\end{figure*}

The extraction of the rPPG signal generally follows unsupervised methods that rely on a structured pipeline \cite{rppg_toolbox}, where regions of interest (ROIs) on the skin are isolated using computer vision techniques \citep{ambient_light,roi_nose,CIELab,cnn+pos,fourpage_pos,Spatial+pos,roi_classic_pos,roi_pos,omit}. Then conventional algorithms are applied to convert the RGB signal into the rPPG signal and estimate the heart rate \cite{chrom,pbv,pos,omit,green1,green2,pulsegan,ica,lgi}. However, over the past decade, deep learning-based approaches have taken over many parts of processing. These deep learning methods combine conventional techniques with deep learning models or provide end-to-end solutions \cite{AutoHR,contrast_end,Contrastive_Learning,cnn_end,deepphys,effphysc,metalearning,physformer,physnet,RhythmNet,ts_can,bigsmall}. In the case of end-to-end deep learning methods, the raw video input is processed through various network architectures to directly output the physiological signal.

Unsupervised methods for recovering physiological signals typically begin by selecting the area of the skin that is most likely to produce high-quality signals, with factors such as skin thickness, movement, and lighting playing a significant role \cite{roi_pos}. Previous studies have shown that regions such as the cheeks and forehead are often reliable sources of extraction of strong rPPG signals, while areas around the mouth and eyes tend to produce noisy data \cite{dynamic_roi,goodbad}. In addition, some research has focused on regions such as the hands \cite{handandface,hand1,hand2} or neck \cite{neckandface,neckandface2,neck1}, which can provide high-quality signals due to the abundance of capillaries and arteries in these areas. However, there is a lack of research on dynamic approaches that utilize multiple skin regions across the face and body, enabling more robust signal extraction. This would reduce reliance on specific areas that can be blocked or compromised due to factors such as facial expressions, occlusions, or challenging lighting conditions \cite{noise_ill_robust,eval_occilation}, ultimately offering a more versatile and reliable approach for extracting rPPG signals.

A key requirement for validating rPPG methods is testing them on realistic datasets. Although existing datasets provide video recordings with ground-truth physiological signals \cite{ubfc,ubfc_phys,pure,mahnob_hci,lgi}, the two are not synchronized. Videos are recorded at a fixed frame rate (FPS), whereas reference signals are sampled at different frequencies, resulting in misaligned timestamps. The interpolation or resampling needed to align them can distort the data. Moreover, existing datasets lack real-world complexity, as they do not include head movements, facial expressions, and variations in lighting. Some are restricted to controlled scenarios with plain, high-contrast backgrounds and depend on high-quality cameras, making them not representative of real-world settings.

We introduce a unified pre-processing model for extracting the rPPG signal. Our model generates a dense mask that identifies skin pixels throughout the body, enabling the segmentation of skin even when insufficient facial regions are visible. Furthermore, it generates a mask that provides a pixel-wise weighting to indicate its contribution to signal extraction. In this way, the model suppresses noisy areas and highlights regions with better conditions for signal extraction. Both segmentation and weighting are integrated into a single model, simplifying the pipeline and enabling real-time performance. Overall, the technique provides robustness under real-world conditions and, through its unsupervised pipeline, guarantees strong generalization across datasets. In addition, we introduce a new dataset with diverse real-life scenarios and synchronized sampling rates, which offers a realistic benchmark for evaluating rPPG methods.

\subsection*{Contributions}
The contribution of this paper falls into two categories:
\begin{itemize}
    \item We propose SkinMap, a DeepLabV3-based model that segments facial and body skin, producing a weighted mask for each frame that prioritizes regions with stronger signal quality, without requiring any additional face or landmark detection.
    \item We present SYNC-rPPG, a new dataset that captures data in four real-world scenarios across 80 samples. Data collection was done using an affordable camera and sensor with the same sampling rates. 
\end{itemize}

\section{Related Work}
\label{sec:related}
In recent years, multiple approaches have been proposed for extracting heart rates from video cameras, ranging from unsupervised to fully end-to-end supervised models. Unsupervised methods typically involve segmentation of the skin region followed by applying conventional techniques, such as LGI \cite{lgi},POS \cite{pos}, CHROM \cite{chrom}, PBV \cite{pbv}, PCA \cite{pca}, OMIT \cite{omit}, GREEN \cite{green1,green2}, to extract the rPPG signal and apply denoising.  
\subsection{Unsupervised Methods}
In facial video-based rPPG, several works have focused on selecting ROIs to achieve the most reliable rPPG signals. \cite{roi_pos} conducted experiments on 39 anatomically divided facial regions to identify the most accurate regions for rPPG extraction, highlighting that the cheeks and forehead are more reliable based on their anatomical characteristics. Similarly, in the work \cite{forhead_cheaks}, the same three main rPPG signal sources (cheeks and forehead) are selected. Many similar studies proposing new unsupervised algorithms use face detection in combination with spatial averaging over the entire skin area as the ROI. In particular, the rPPG-toolbox \cite{rppg_toolbox}, a comprehensive toolbox for rPPG signal processing, utilizes spatial averaging for each frame in an unsupervised pipeline. Face2PPG \cite{omit} has been introduced to stabilize movement and expressions using rigid mesh normalization to extract consistent RGB signals, and combines dynamic multi-region selection with OMIT techniques for accurate heart rate estimation. However, it depends on skin detection and facial segmentation and is limited to facial regions. 

As suggested by \cite{segnet}, excluding active areas such as the eyelids and lips helps mitigate motion artifacts, while glasses and hair can contaminate the signal. Narrowing the face area or dividing it into smaller sections without proper skin segmentation increases sensitivity to noise. These issues can be addressed by accurately segmenting the largest possible skin area to improve signal reliability. \cite{fourpage_pos} provides three methods for skin segmentation: two classical color thresholding approaches (Cheref \cite{cheref}, Levelset \cite{levelset}) and a model (DeepLabV3+ \cite{deeplab}) to create a valid face skin mask for the extraction of the rPPG signal. Another interesting study \cite{hand1} proposes the use of rPPG signals to prevent spoofing in palm images by converting RGB frames to YCbCr, with skin pixels identified in the Cb-Cr plane. In \cite{neckandface}, both the neck and the face are used as ROIs to extract the rPPG signal. 

Many rPPG signal extraction algorithms rely on a well-defined, dynamic, weighted skin mask to improve rPPG signal reliability and robustness, and spatially-based techniques often prove to be effective \cite{Spatial+pos}. \cite{review_skinsegmentation} reviewed the past decade of skin segmentation techniques, including deep and non-deep learning approaches. Many studies use MediaPipe's 3D face mesh for ROI extraction \cite{mediapipe_landmark}. The MediaPipe multi-class selfie segmentation model detects face and body skin in real-time \cite{MediaPipe}. However, it does not differentiate between non-skin areas, such as the eyes, mouth, or glasses, and there is limited published work on this model. 

\subsection{Supervised Methods}
 Deep Neural Networks, particularly Convolutional Neural Networks, have gained significant attention in computer vision and signal processing, including healthcare applications. An end-to-end model directly maps raw video frames to physiological signals, requiring dataset-specific training with ground-truth rPPG signals so that the network learns the entire extraction process without a chain of pre-processing steps such as face detection, skin segmentation, color space transformation, or signal filtering.
 
 DeepPhys \cite{deepphys} is an end-to-end convolutional attention network that estimates heart rate and breathing rate directly from video. The method introduces a motion representation based on normalized frame differences. It uses an appearance-guided attention mechanism that learns soft masks to highlight informative skin regions. EfficientPhys \cite{effphysc} introduces a convolution-based network with a custom normalization module (difference + batchnorm), tensor-shifted convolutions, and self-attention for efficient spatiotemporal modeling. PhysFormer \cite{physformer} is also an end-to-end video transformer. It introduces temporal difference transformer blocks that combine temporal-difference multi-head self-attention (TD-MHSA) and spatio-temporal feed-forward (ST-FF) modules. 

Extraction of rPPG signal relies on extremely subtle, quasi-periodic pixel changes caused by blood volume fluctuations in the skin, which are easily overshadowed by much larger variations from factors such as lighting conditions and motion. Unsupervised approaches tend to offer better generalization in different applications \cite{Style-rPPG}. Although attention mechanisms are powerful in computer vision tasks with rich spatial semantics, in rPPG, they often amplify dataset-specific textures, lighting artifacts, or camera noise  \cite{deep_end}. This not only results in a lack of understanding of the underlying physiological mechanisms, but also introduces substantial computational overhead \cite{cnn+pos}. In contrast, pre-processing models can attenuate illumination and motion artifacts without requiring the use of the signal itself. The pipeline does not require heavy attention modules to learn which pixels to trust based on the signal. If the pre-processing model enforces physiology-driven priors, it makes the extracted features less biased toward dataset-specific appearances.

\section{Methodology}

\label{sec:metho}

As shown in \cref{fig:pipeline}, the unsupervised pipeline for extracting rPPG signals typically involves the following steps:

\begin{enumerate}
\item Dataset Collection: This step involves collecting video data synchronized with a reference signal and providing the necessary information to read and manage the available or collected dataset.

\item Video Processing: A skin segmentation or ROIs selection technique is applied, followed by average or weighted averaging of the pixel values within the skin region to obtain the RGB signal throughout the video.

\item RGB to rPPG conversion: Transforming skin color variations into physiological signals using algorithms that combine the RGB channels, band-pass filtering, and de-noising to extract the rPPG signal.

\item Heart Rate Estimation: Heart rate is estimated by performing a frequency analysis on the rPPG signal.

\item Evaluation of results: The estimated Heart rate is evaluated based on various metrics to assess its accuracy, robustness, and reliability.
\end{enumerate}

\subsection{Preliminary}

In this work, our objective is to develop a robust skin segmentation framework with a pixel-wise weighting mask to improve the extraction of PPG signals from video. For the skin segmentation task, we adopt a variant of the well-established DeepLabV3 architecture with a ResNet-50 backbone, chosen because it represents a state-of-the-art solution for semantic segmentation while remaining computationally efficient. DeepLabV3 incorporates dilated convolutions and an Atrous Spatial Pyramid Pooling (ASPP) module, which together balance fine spatial detail, which is essential for generating accurate pixel-level masks, with contextual understanding that helps separate skin from background under challenging conditions. Since rPPG depends on pixel averaging rather than edge precision, we only require reliable skin separation with pixel-wise weighting. DeepLabV3 provides this balance efficiently, while heavier models add unnecessary complexity without clear benefit.

\begin{figure}[t]
  \centering
   \includegraphics[width=1\linewidth]{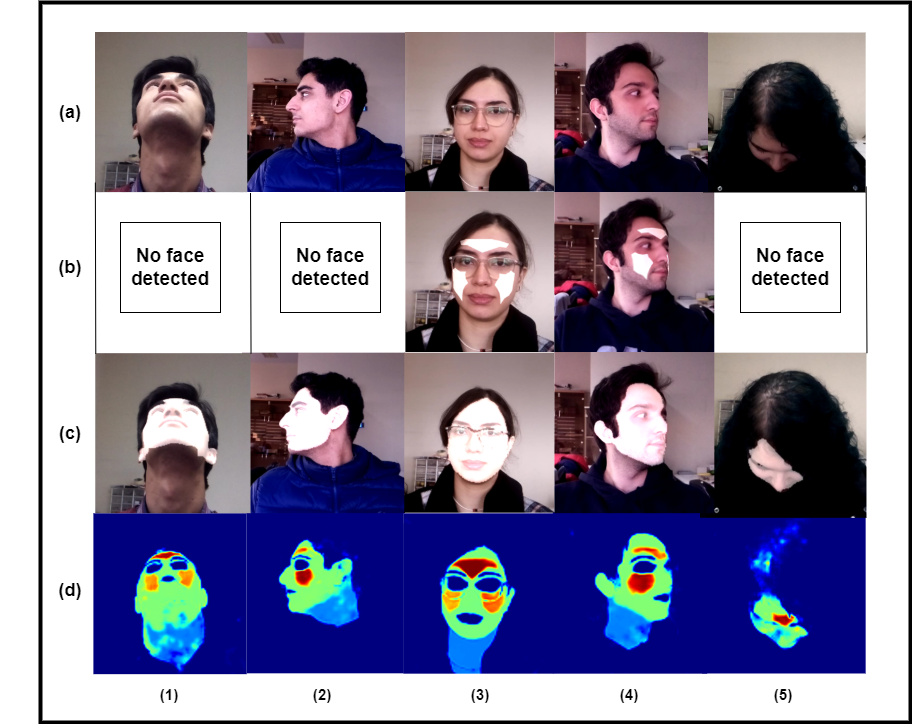}

   \caption{Illustration of our dataset and segmentation results.  
(a) Frame samples from the rotation task of our dataset.
(b) Segmentation results using Face Landmark Detection, where white areas indicate detected ROIs. In some frames, the Landmarker failed to detect a face.
(c) Segmentation results using the Multi-Class Selfie Segmentation, where white areas represent detected skin regions.  
(d) Heat-map visualization of the output of our segmentation model.}
\label{fig:segmentation1}

\end{figure}

In the process of training and evaluating our model, we use two state-of-the-art MediaPipe skin segmentation methods. The first is Face Landmark Detection of MediaPipe \cite{mediapipe_landmark}. A real-time model that predicts 468 3D facial landmarks, employing BlazeFace face detection followed by 3D landmark regression using a MobileNetV2 backbone optimized through transfer learning and Euclidean loss minimization. It can be used to determine the pulse of the cheeks and forehead, which are widely used as ROIs in rPPG extraction pipelines \cite{roi_pos}. The second one is Multi-Class Selfie Segmentation of MediaPipe (MCSS). A Vision Transformer-based model designed for real-time segmentation of human subjects. It outputs segmentation masks at 256×256×6 and 512×512×6 resolutions, including background, hair, body skin, face skin, clothing, and accessories classifications. However, it does not explicitly differentiate non-skin facial areas, such as the eyes, mouth, or glasses \cite{MediaPipe}. It is worth noting that all three models are strong performers and capable of real-time processing. A comparison of the results from MediaPipe Landmarker, Multi-Class Selfie Segmentation, and our trained model is illustrated in \cref{fig:segmentation1}. As shown, the Landmarker failed to detect the face at harsh angles and when it was not fully visible.

\subsection{Proposed Architecture}

Traditional algorithms average pixel values from selected areas of facial skin. However, an intelligent system is needed to automatically segment skin regions, compute their average pixel values, and feed them into the pulse extraction algorithm. As mentioned, we are using the DeepLabV3-ResNet50 architecture \cite{deeplabv3}. We replace the final layer of the default and auxiliary classifier with a single-channel convolutional layer. A sigmoid activation function is attached to the final layer to confine the output values between 0 and 1. After fine-tuning the model on a large and suitable dataset, we expect it to effectively segment all available skin areas by generating a mask, assigning weights based on the subject’s position and lighting conditions in each frame of the video dataset. 

\subsection{Photo Dataset Creation}

Training our segmentation model requires a diverse dataset of human images under various lighting and environmental conditions. Although there are some public skin segmentation datasets \cite{photodataset1,photodataset2,photodataset3}, they contain a very limited number of precise samples and are not suitable for our training \cite{review_skinsegmentation}. To address this, we curated a custom dataset by extracting human images from the COCO dataset \cite{cocodataset}, which offers a rich variety of real-world scenes with diverse backgrounds. 

In the first step of generating our training dataset, we need to extract images containing humans with fully visible faces. To filter the dataset, we utilize the MediaPipe Face Landmark Detector to identify and select relevant images, which feature individuals and groups in various age ranges. Secondly, we need to generate a reference skin mask for each of these extracted images to train our model later.

As discussed, various regions of the face influence signal quality unequally. The lips and mouth have a different color from the rest of the skin and the amplitude of the heart pulse extracted from these areas is negligible. The eye region is prone to excessive movement, which introduces noise into the signal \cite{roi_pos,segnet}. Previous studies have shown that the cheeks and forehead exhibit the highest amplitude of the pulse signal \cite{roi_pos}. Therefore, prioritizing the segmentation of these regions over other skin areas is essential for accurate signal extraction. For the synthesized photo dataset, we need to systematically assign importance to different facial regions. We classify them into three priority levels:  

\begin{itemize}
    \item Priority 1: Forehead and cheeks, as they provide the highest-quality pulse signals.
    \item Priority 2: Other facial skin regions, excluding areas around the eyes, eyebrows, and lips.
    \item Priority 3: Other skin surfaces on the body.
\end{itemize}

Regions with higher priority should have greater weight in the final skin mask. To achieve this, we introduce a weighting mechanism that considers both the angular orientation of the skin relative to the camera and the assigned priority level. For Priority 1 regions, the weight varies between 4 and 2 depending on the angle, while Priority 2 and 3 regions are assigned fixed weights of 2 and 1, respectively. For example, suppose that the subject is looking directly at the camera; in this case, the specular reflection of the forehead is maximized \cite{angle}. We use facial landmarks to estimate the orientation of each region. The weighting function for Priority 1 regions is defined in equation \cref{eq:weighting}. To explain \cref{eq:weighting}, we assign a weight \( P_i \) to each region based on the angle between the normal vector of the surface and the direction of the camera \( \theta_i \). We adjusted the ROIs weighting so that smaller angles receive a higher weight. This results in a weight curve ranging from 2 to 4. The cosine function, used for the effective area, smooths the curve and minimizes noise in challenging poses. 

\begin{equation}
P_i = 2\bigl(\cos(\theta_i) + 1\bigr), \quad \theta_i \leq \tfrac{\pi}{2}
\label{eq:weighting}
\end{equation}

To construct the skin mask for our photo dataset, we first use the MediaPipe Selfie Segmentation model to extract the face and body skin (assigned as a priority 2 and 3 region). Next, we utilize the MediaPipe Face Landmarker to exclude the eyes, eyebrows, and mouth, thereby minimizing noise and also defining priority-1 regions. We then combine these outputs and assign weights based on \cref{eq:weighting} and the priorities. After these steps, the mask is normalized to a value between 0 and 1 to maintain consistency with the network output scale. This process is applied to selected human images from the COCO dataset, producing an image–skin mask dataset of 8,000 samples with reliable ground-truth masks for skin segmentation and weighting. This dataset is subsequently used to train our DeepLabV3 model for accurate and weighted skin segmentation.

\begin{figure}[t]
  \centering
   \includegraphics[width=1\linewidth]{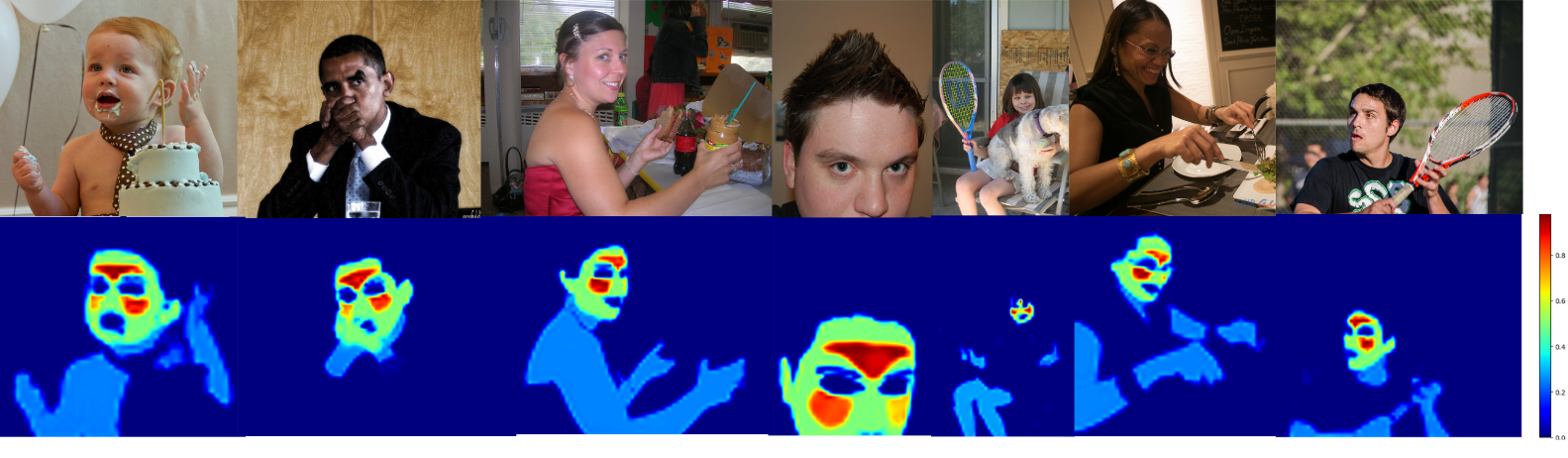}

   \caption{Model output on a random sample from the COCO \cite{cocodataset} dataset, showcasing its reliability in real-world applications.}

\label{fig:coco}

\end{figure}

We should emphasize that the MediaPipe models are used only for creating the photo dataset. Furthermore, our experiments demonstrate that the final trained model surpasses all of these baseline models in performance while remaining computationally as efficient as they are. \cref{fig:coco} presents sample output of the trained model on randomly selected images from the COCO dataset \cite{cocodataset}. These results demonstrate the robustness of the model to skin tone variations.

\subsection{Model Training and Heart Rate Estimation}
\label{subsec:training}

We trained our DeepLabV3-ResNet50 model for 30 epochs, using 90 percent of the data for training and 10 percent for validation. The training was conducted on an RTX 4090 GPU with 20GB of VRAM usage, supported by 198GB DDR5 RAM and an Intel i7-14700K CPU. The training took approximately 4 hours. During training, the training loss steadily decreased and converged, and the validation loss, despite initial fluctuations, trended downward. We stopped at epoch 30, ensuring effective learning and generalization. For each video in the rPPg dataset, a weighted average of the pixels is computed for every frame based on the model’s output. From this, the corresponding RGB signal is generated. Based on the implementation of \cite{rppg_toolbox}, the RGB signal of each video is processed using commonly used rPPG algorithms. The extracted rPPG signal is then used to estimate heart rate (HR). Heart rate is determined using the Fourier transform (FFT), and band-pass filtering, which extracts frequency components within the physiological heart rate range. The strongest frequency in this range is identified as the heart rate in beats per minute (BPM).

\section{Experiments}
\label{sec:results}

We evaluated the proposed segmentation model in terms of both accuracy and robustness of rPPG signal extraction by comparing it to state-of-the-art unsupervised and supervised settings. We incorporate several pre-processing techniques into our comparison, including the rPPG-Toolbox preprocessing based on spatial averaging \cite{rppg_toolbox}, Mediapipe Landmark Detection to isolate the cheek and forehead areas with equal weighting, and Mediapipe Multi-Class Selfie Segmentation (MCSS) to detect facial skin regions. In addition, we include the non-weighted version of our model, which performs full-body skin segmentation, to highlight the benefits of intelligent weighting and the inclusion of body skin. We refer to this model as the full-body model in this section. We mainly employ the POS algorithm \cite{pos} for unsupervised pipelines due to its proven superior performance compared to other methods \cite{forhead_cheaks}. In the supervised pipeline, we report the results of the pre-trained models of DeepPhys \cite{deepphys}, EfficientPhys \cite{effphysc}, and PhysFormer \cite{physformer}. 

The comparison is conducted on our dataset as well as the UBFC-PHYS dataset \cite{ubfc_phys}. These datasets are selected because they provide sufficient subject diversity and include several real-world scenarios. Since our goal is to evaluate the reliability of these techniques, we focus on identifying the best model that can generalize across scenarios and subjects. To ensure fairness, we utilized the pre-trained versions of the supervised models, which were originally trained on the UBFC-rPPG dataset \cite{ubfc}. This dataset represents a relatively simple setting with stationary subjects and ideal lighting conditions. In contrast, UBFC-PHYS and SYNC-rPPG are significantly more complex.

\subsection{Experimental Setup}

In this paper, we present a new rPPG dataset. Most available datasets are captured under ideal lighting and environmental conditions, with minimal subject movement, which does not accurately represent real-life applications. Based on these circumstances, we consider it essential to design and implement a dedicated sampling device that ensures the precise, simultaneous acquisition of image and pulse data. An overview of our setup is provided in \cref{fig:pipeline} (a). We select a Raspberry Pi 4B development board featuring a 64-bit processor clocked at 1.5 GHz and 8 GB of RAM. The system runs the Raspberry Pi OS and utilizes Python for rapid development and seamless integration. For video capture, we employ the Raspberry Pi Camera V2 module, which provides imaging at a resolution of 1280×720 pixels and a frame rate of 30 frames per second (fps). For the data collection, we integrate a laboratory-grade MAX30102 sensor to capture heart pulse data. Furthermore, to improve measurement reliability and reduce errors in pulse signal capture, an additional MAX30102 sensor is integrated to simultaneously acquire pulse data from both hands.

The sensors and camera are precisely synchronized at 30 FPS, which is the maximum achievable rate limited by the intrinsic capabilities of the camera module, thus ensuring a stable high-rate data stream for accurate rPPG analysis. As demonstrated in \cite{omit}, the PPG signals from fingertip contact-based sensors in publicly available datasets exhibit fluctuations due to finger movement or disconnections, resulting in errors in heart rate estimation. Since we collected the data ourselves, we know the challenges of working with associated devices. SYNC-rPPG incorporates two sensors, and we use the mean value of their signals.

\subsection{Datasets}

Our dataset, named SYNC-rPPG, was collected from 20 individuals, with each video lasting 30 seconds. All subjects gave their informed consent for their data to be made publicly available. Each participant was recorded in four different scenarios. In the first scenario, the subject remained calm with no head movement and minimal facial expressions. In the second scenario, the subject was asked to read an emotional passage or discuss an important personal memory. In the third scenario, the subject performed rapid head rotations. In the fourth scenario, the recording took place after exercise, under conditions similar to the first scenario. A comparison between SYNC-rPPG and other datasets used in this study is presented in \cref{tab:dataset_comparison}. 

\begin{table}
\resizebox{0.5\textwidth}{!}{ % Fit within half-page width
\begin{tabular}{l|c|c|c}
\textbf{Attribute} & \textbf{UBFC-rPPG} & \textbf{UBFC-PHYS} & \textbf{SYNC-rPPG} \\
\hline
Sample count     & 50 & 168 & 80 \\
Scenarios            & rest & rest, talk, exercise & rest, talk, rotation, exercise \\
Video (FPS)       & 30 & 35 & 30 \\
Sensor (Hz)      & 60 & 64 & 30 \\
Resolution           & 640×480 & 1024×1024 & 1280×720 \\
Heart rate Range (bpm) & $\sim$60--80 & $\sim$60--100 & $\sim$60--140 \\
Lighting             & perfect & perfect & day-light + artificial \\
Sensor count    & 1 & 1 & 2 \\

\end{tabular}
}
\caption{Comparison of rPPG Datasets}
\label{tab:dataset_comparison}
\end{table}

The UBFC-RPPG database \cite{ubfc} utilizes a Logitech C920 HD Pro webcam at 30 frames per second (fps) and 640x480 resolution in uncompressed 8-bit RGB format. A CMS50E pulse oximeter is used to capture PPG data. The database includes 50 videos, each approximately 1 minute long and featuring minimal movement. The UBFC-PHYS dataset \cite{ubfc_phys} includes data from 56 subjects, participating in three tasks: rest, speech, and arithmetic. Participants are filmed and wear a wristband that records BVP and EDA signals. 

\subsection{Experimental Results}

To evaluate the extracted heart rate, we employ five metrics: mean absolute error (MAE), root mean square error (RMSE), mean absolute percentage error (MAPE) as introduced in \cite{segnet}, Pearson correlation coefficient (PCC), and signal-to-noise ratio (SNR). The runtime performance of our model on SYNC-rPPG achieved an average processing speed of 211.85 FPS with an average latency of 6.65 ms on an NVIDIA RTX 3060 GPU, demonstrating that the model is capable of real-time operation.

\begin{figure}
        \centering
        \includegraphics[width=\linewidth]{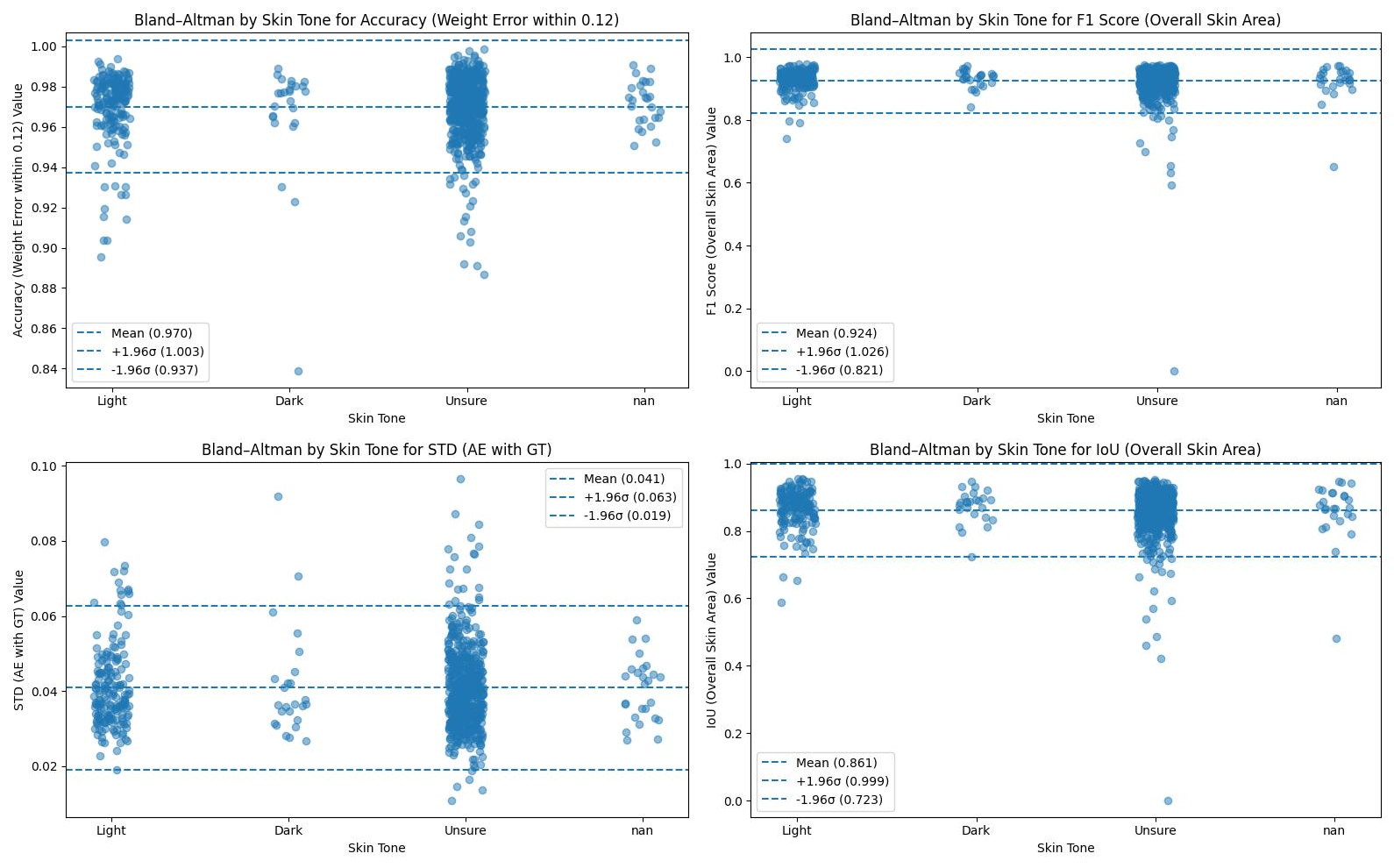}
        \caption{Evaluating skin segmentation by skin tone. Top left: accuracy (Weight Error within 0.12). Top right: F1 score (Overall Skin Area). Bottom left: standard deviation (AE with GT). Bottom right: IoU (Overall Skin Area)}
        \label{fig:f1_score_overall_skin}
\end{figure}

This work focuses on improving the rPPG signal extraction, rather than introducing the most powerful segmentation model; however, we evaluate segmentation accuracy to ensure reliability. We analyze the accuracy and diversity of our model, with and without weights, across different skin tones using the annotations (light, dark, unsure, and nan) provided in \cite{r4} for the COCO human image dataset. For validation, we used 10 percent of our synthesized dataset. The weighted mask achieves a mean accuracy of 0.97 and a mean F1 score of 0.924 for overall skin detection, as shown in \cref{fig:f1_score_overall_skin}. These results indicate that the model is capable of highly accurate and consistent skin segmentation across diverse skin tones, demonstrating strong generalization to population diversity.

\begin{figure}
            \centering
            \includegraphics[width=\linewidth]{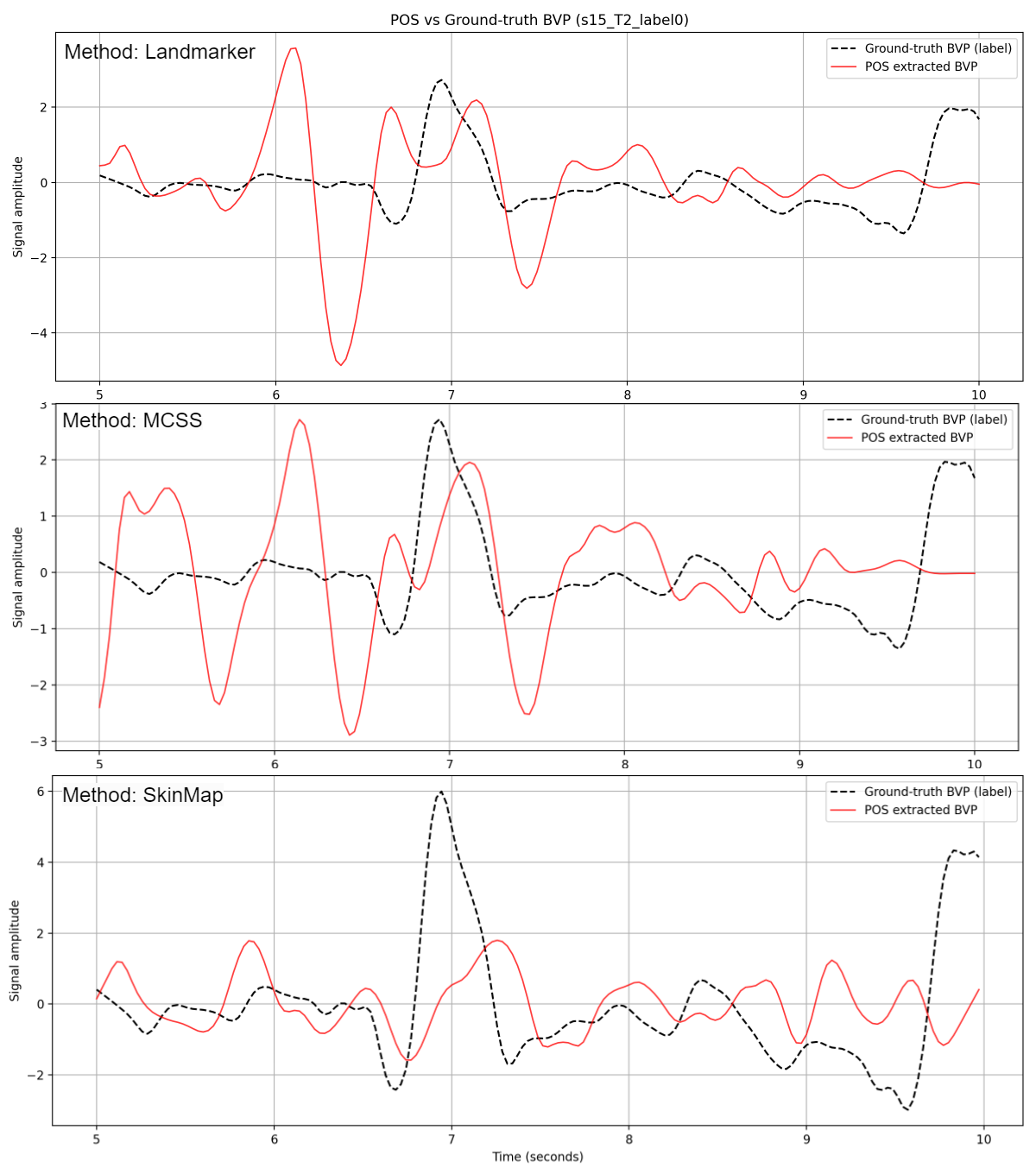}
            \caption{Extracted signals using models, up: Face Landmark Detection, middle: Multi-Class Selfie Segmentation, down: SkinMap.}
   \label{fig:plots}
\end{figure}

One way to evaluate models is by measuring the average number of frames in which they fail to adjust a mask and therefore cannot contribute to the final RGB signal. This problem is more common in ROIs-based models during motion, as shown in \cref{fig:segmentation1}, where they do not detect the correct region or the face detector could not locate the face. In our dataset, Face Landmark Detection misses an average of 0.75 frames per video in talking tasks and 118 frames per video in rotation tasks, whereas SkinMap and Multi-Class Selfie Segmentation perform flawlessly. \cref{fig:plots} are extracted signals from one of the samples of UBFC-Phys dataset in the talking scenario. It is evident that our model can reconstruct the true shape and peaks of the signal much more accurately.

\begin{figure}[t]
    \centering
    \includegraphics[width=1\linewidth]{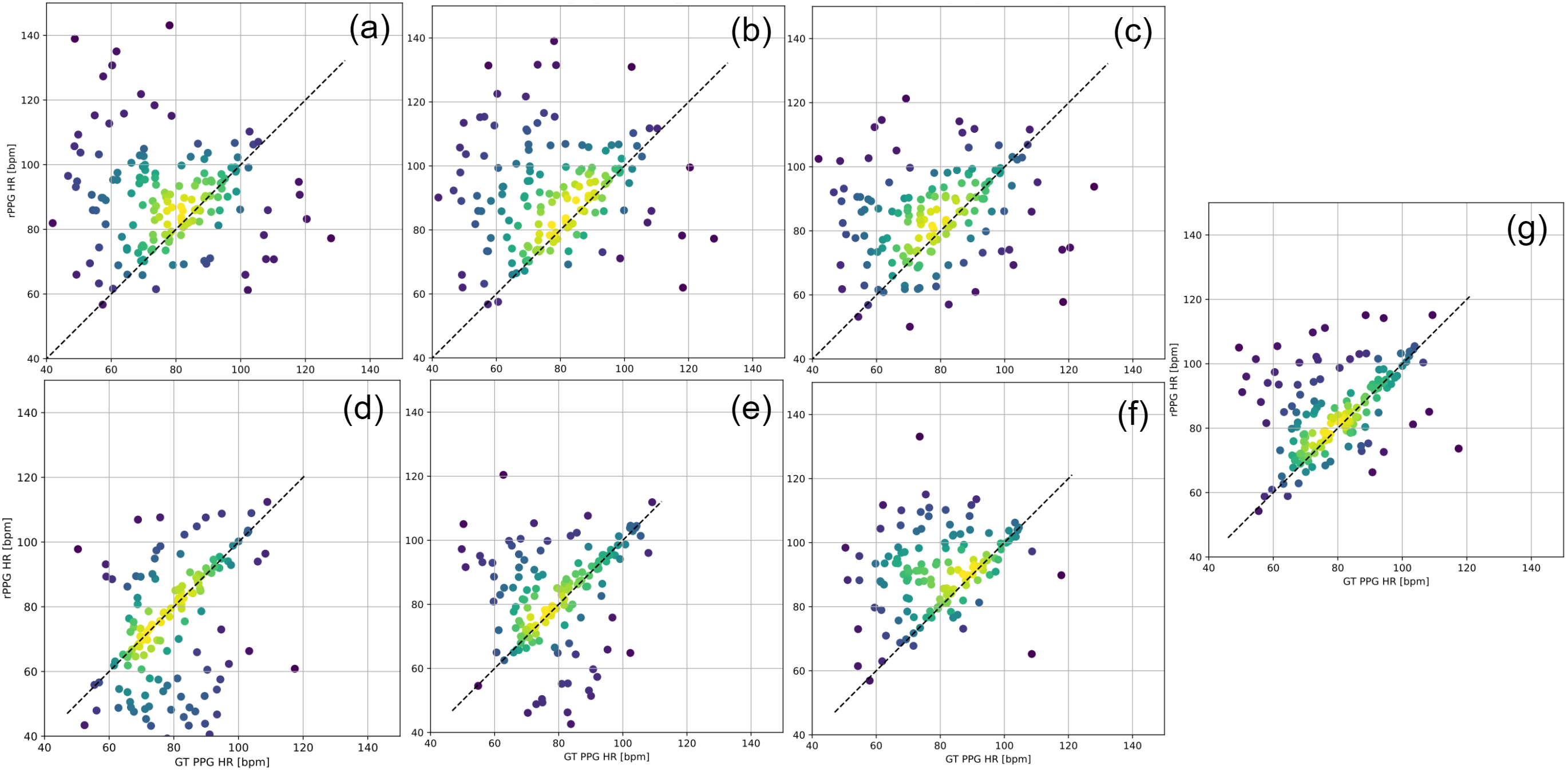}
    \caption{Predicted vs. ground-truth haert rate (BPM) on the UBFC-Phys: (a) Landmarker, (b) MCSS, (c) Full-body, (d) DeepPhys, (e) EfficientPhys, (f) PhysFormer, (g) SkinMap. The dashed line shows perfect prediction.}
    \label{fig:ubfcphys_scater}
\end{figure}

\begin{figure}[ht]
        \centering
        \includegraphics[width=\linewidth]{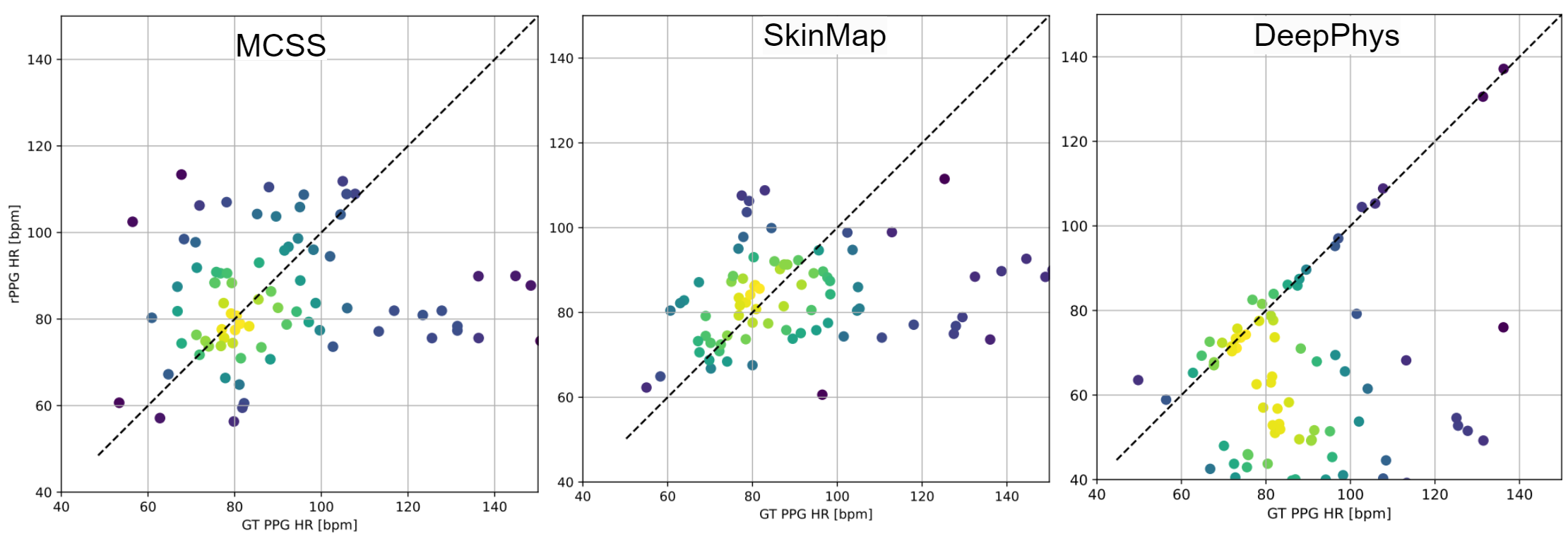}
        \caption{Predicted vs. ground-truth heart rate (BPM) scatter plot on the SYNC-rPPG dataset. Left: MCSS, middle: SkinMap, and right: DeepPhys.}
        \label{fig:scatersync}
\end{figure}

As illustrated in \cref{fig:ubfcphys_scater}, plots (a) to (c) and (g) present the pre-processing results of unsupervised pipelines, while panels (d) to (f) correspond to the supervised settings on the UBFC-Phys dataset. The results indicate that SkinMap achieves the tightest clustering along the diagonal, reflecting more consistent and accurate predictions. Furthermore, SkinMap demonstrates superior and more generalized performance compared to pre-trained supervised models. This finding suggests that supervised models trained on simplified settings fail to outperform an unsupervised pipeline equipped with SkinMap, highlighting their limited reliability in healthcare applications. \cref{fig:scatersync} presents the results on the SYNC-rPPG dataset; SkinMap pipeline achieves the lowest variance around the diagonal. In addition, we observed that rPPG extraction algorithms, such as POS, struggle to reconstruct signals at higher heart rates, while supervised approaches handle them better; however, their performance is highly dependent on the training data. 

    \begin{table*}[ht!]
\centering
\scriptsize
\begin{adjustbox}{max width=\textwidth}
\begin{tabular}{@{}lllccccccc|c@{}}
\toprule
& & & \multicolumn{8}{c}{\textbf{Models}} \\
\cmidrule(lr){4-11}
\textbf{Dataset} & \textbf{Scenario} & \textbf{Metric} & Spatial Average \cite{rppg_toolbox} & Landmarker & MCSS & Full-body & DaapPhys \cite{deepphys} & EfficientPhys \cite{effphysc} & PhysFormer \cite{physformer} & \textbf{SkinMap (ours)} \\
\midrule

% ---------------- SYNC-rPPG ----------------
\multirow{20}{*}{SYNC-rPPG} 
& \multirow{5}{*}{Rest} 
  & MAE~$\downarrow$    & 11.34$\pm$2.48 & 10.02$\pm$2.81 & 11.51$\pm$3.16 & 10.55$\pm$2.15 & 5.89$\pm$2.97 & \textbf{1.85$\pm$0.61} & 11.87$\pm$2.48* & 6.86$\pm$1.62 \\
& & RMSE~$\downarrow$   & 15.86$\pm$9.63 & 16.08$\pm$11.37 & 18.24$\pm$11.91 & 14.26$\pm$8.60 & 14.54$\pm$12.31 & 3.31$\pm$2.27 & 16.24$\pm$9.54 & 9.97$\pm$5.54 \\
& & MAPE~$\downarrow$   & 15.41$\pm$3.59 & 13.92$\pm$4.51 & 16.41$\pm$4.93 & 14.72$\pm$3.44 & 6.98$\pm$3.33 & 2.39$\pm$0.78 & 16.49$\pm$3.89 & 9.09$\pm$2.15 \\
& & PCC~$\uparrow$  & -0.04$\pm$0.24 & 0.11$\pm$0.23 & 0.20$\pm$0.23 & 0.039$\pm$0.236 & 0.233$\pm$0.229 & 0.933$\pm$0.085 & -0.042$\pm$0.235 & 0.497$\pm$0.205 \\
& & SNR (dB)~$\uparrow$ & -5.49$\pm$0.45 & -4.82$\pm$0.45 & -4.75$\pm$0.43 & -4.74$\pm$0.52 & -1.66$\pm$0.74 & -1.44$\pm$0.64 & -5.21$\pm$0.43 & -4.20$\pm$0.51 \\
\cmidrule(lr){2-11}

& \multirow{5}{*}{Talking}
  & MAE~$\downarrow$    & 13.45$\pm$2.39 & 13.54$\pm$2.59 & 12.83$\pm$2.40 & \textbf{12.30$\pm$2.04} & 29.53$\pm$4.44* & 22.85$\pm$4.28 & 12.39$\pm$2.55 & 12.66$\pm$2.17 \\
& & RMSE~$\downarrow$   & 17.16$\pm$9.50 & 17.81$\pm$9.92 & 16.74$\pm$9.31 & 15.31$\pm$8.63 & 35.59$\pm$16.91 & 29.81$\pm$15.64 & 16.85$\pm$9.28 & 15.95$\pm$9.33 \\
& & MAPE~$\downarrow$   & 15.44$\pm$2.72 & 14.86$\pm$2.53 & 14.35$\pm$2.52 & 13.75$\pm$2.11 & 31.41$\pm$4.38 & 24.65$\pm$4.26 & 13.96$\pm$2.97 & 14.42$\pm$2.31 \\
& & PCC~$\uparrow$ & 0.243$\pm$0.229 & 0.31$\pm$0.22 & 0.32$\pm$0.22 & 0.439$\pm$0.212 & -0.281$\pm$0.226 & -0.262$\pm$0.227 & 0.128$\pm$0.234 & 0.242$\pm$0.229 \\
& & SNR (dB)~$\uparrow$ & -6.57$\pm$0.51 & -6.09$\pm$0.56 & -6.34$\pm$0.64 & -6.21$\pm$0.59 & -8.35$\pm$0.77 & -7.24$\pm$0.70 & -5.95$\pm$0.34 & -6.15$\pm$0.67 \\
\cmidrule(lr){2-11}

& \multirow{5}{*}{Head Rotation}
  & MAE~$\downarrow$    & 14.85$\pm$2.10 & 24.17$\pm$3.51 & 13.80$\pm$1.92 & 13.45$\pm$2.47 & 27.25$\pm$2.36* & 21.45$\pm$3.18 & 15.21$\pm$2.71 & \textbf{11.95$\pm$2.13} \\
& & RMSE~$\downarrow$   & 17.58$\pm$8.32 & 28.82$\pm$14.04 & 16.25$\pm$7.91 & 17.41$\pm$9.88 & 29.22$\pm$11.35 & 25.74$\pm$12.80 & 19.44$\pm$10.63 & 15.29$\pm$8.17 \\
& & MAPE~$\downarrow$   & 19.18$\pm$2.93 & 31.30$\pm$4.39 & 17.74$\pm$2.57 & 17.65$\pm$3.40 & 34.05$\pm$2.51 & 27.80$\pm$4.44 & 20.50$\pm$4.24 & 14.99$\pm$2.59 \\
& & PCC~$\uparrow$  & -0.028$\pm$0.236 & 0.50$\pm$0.20 & 0.03$\pm$0.24 & 0.170$\pm$0.232 & -0.072$\pm$0.235 & -0.335$\pm$0.222 & 0.107$\pm$0.234 & 0.343$\pm$0.221 \\
& & SNR (dB)~$\uparrow$ & -6.25$\pm$0.38 & unstable & -6.90$\pm$0.47 & -5.62$\pm$0.44 & -9.35$\pm$0.60 & -7.77$\pm$0.48 & -6.07$\pm$0.41 & -5.82$\pm$0.49 \\
\cmidrule(lr){2-11}

& \multirow{5}{*}{After Exercise}
  & MAE~$\downarrow$    & 36.47$\pm$4.86 & 29.53$\pm$5.33 & 32.70$\pm$5.27 & 33.05$\pm$4.89 & 45.18$\pm$9.34* & 37.88$\pm$7.69 & \textbf{28.56$\pm$5.67} & 32.96$\pm$4.64 \\
& & RMSE~$\downarrow$   & 42.46$\pm$17.89 & 37.96$\pm$18.98 & 40.31$\pm$19.11 & 39.62$\pm$18.05 & 61.53$\pm$31.25 & 51.16$\pm$25.13 & 38.21$\pm$20.06 & 38.94$\pm$17.70 \\
& & MAPE~$\downarrow$   & 29.05$\pm$3.38 & 22.95$\pm$3.80 & 25.77$\pm$3.54 & 25.82$\pm$3.29 & 34.82$\pm$6.73 & 28.44$\pm$5.49 & 21.98$\pm$3.95 & 26.11$\pm$3.07 \\
& & PCC~$\uparrow$  & 0.241$\pm$0.229 & 0.00$\pm$0.24 & -0.47$\pm$0.21 & 0.033$\pm$0.236 & -0.317$\pm$0.224 & -0.450$\pm$0.210 & -0.038$\pm$0.236 & 0.312$\pm$0.224 \\
& & SNR (dB)~$\uparrow$ & -10.77$\pm$0.95 & -9.64$\pm$1.03 & -10.18$\pm$1.05 & -10.02$\pm$0.83 & -8.65$\pm$1.08 & -7.34$\pm$0.92 & -8.92$\pm$1.03 & -9.49$\pm$0.88 \\
\midrule

% ---------------- UBFC-Phys ----------------
\multirow{15}{*}{UBFC-Phys} 
& \multirow{5}{*}{Rest}
  & MAE~$\downarrow$    & 4.91$\pm$1.23 & 5.13$\pm$1.55 & 5.28$\pm$1.52 & 4.65$\pm$1.10 & 5.57$\pm$1.43 & \textbf{3.75$\pm$0.98} & 6.25$\pm$1.46* & 5.18$\pm$1.36 \\
& & RMSE~$\downarrow$   & 10.13$\pm$6.50 & 12.29$\pm$8.53 & 12.19$\pm$8.52 & 9.19$\pm$5.72 & 11.20$\pm$6.57 & 7.63$\pm$5.17 & 11.87$\pm$7.12 & 10.86$\pm$6.95 \\
& & MAPE~$\downarrow$   & 6.88$\pm$1.89 & 6.83$\pm$2.43 & 7.03$\pm$2.41 & 5.98$\pm$1.64 & 7.47$\pm$2.02 & 5.34$\pm$1.52 & 8.97$\pm$2.27 & 7.27$\pm$2.07 \\
& & PCC~$\uparrow$  & 0.751$\pm$0.093 & 0.577$\pm$0.116 & 0.597$\pm$0.113 & 0.770$\pm$0.090 & 0.718$\pm$0.105 & 0.834$\pm$0.083 & 0.678$\pm$0.108 & 0.717$\pm$0.102 \\
& & SNR (dB)~$\uparrow$ & 0.69$\pm$0.71 & 2.82$\pm$0.90 & 3.06$\pm$0.87 & 2.04$\pm$0.91 & 0.322$\pm$0.771 & 0.71$\pm$0.75 & -0.75$\pm$0.84 & 0.37$\pm$0.80 \\
\cmidrule(lr){2-11}

& \multirow{5}{*}{Talking}
  & MAE~$\downarrow$    & 12.75$\pm$1.80 & 25.00$\pm$2.72* & 24.85$\pm$2.80 & 16.09$\pm$2.01 & 19.45$\pm$2.37 & 16.91$\pm$2.11 & 18.19$\pm$1.95 & \textbf{12.04$\pm$1.73} \\
& & RMSE~$\downarrow$   & 18.20$\pm$9.03 & 31.77$\pm$14.53 & 32.03$\pm$13.29 & 21.64$\pm$10.36 & 25.10$\pm$10.69 & 22.46$\pm$10.25 & 21.98$\pm$9.48 & 17.35$\pm$8.90 \\
& & MAPE~$\downarrow$   & 18.38$\pm$3.07 & 35.87$\pm$4.80 & 35.38$\pm$4.44 & 22.24$\pm$2.90 & 24.31$\pm$2.94 & 23.51$\pm$3.42 & 25.33$\pm$3.21 & 16.82$\pm$2.88 \\
& & PCC~$\uparrow$  & 0.143$\pm$0.140 & -0.262$\pm$0.136 & -0.073$\pm$0.141 & 0.193$\pm$0.139 & -0.062$\pm$0.152 & -0.126$\pm$0.145 & 0.214$\pm$0.158 & 0.124$\pm$0.140 \\
& & SNR (dB)~$\uparrow$ & -5.14$\pm$0.41 & -7.42$\pm$0.54 & -6.18$\pm$0.57 & -6.30$\pm$0.54 & -6.14$\pm$0.43 & -5.53$\pm$0.43 & -6.24$\pm$0.38 & -5.19$\pm$0.40 \\
\cmidrule(lr){2-11}

& \multirow{5}{*}{Arithmetic}
  & MAE~$\downarrow$    & 10.31$\pm$1.62 & 22.13$\pm$2.51* & 20.51$\pm$2.27 & 19.89$\pm$2.08 & 13.18$\pm$1.87 & 12.19$\pm$1.89 & 16.44$\pm$2.12 & \textbf{10.12$\pm$1.61} \\
& & RMSE~$\downarrow$   & 15.68$\pm$7.74 & 28.72$\pm$12.60 & 26.34$\pm$11.57 & 24.99$\pm$10.63 & 18.68$\pm$9.31 & 17.99$\pm$8.39 & 21.94$\pm$10.02 & 15.46$\pm$7.91 \\
& & MAPE~$\downarrow$   & 15.06$\pm$2.70 & 35.88$\pm$4.84 & 33.45$\pm$4.56 & 31.37$\pm$4.21 & 16.86$\pm$2.26 & 17.56$\pm$3.00 & 23.29$\pm$3.24 & 14.72$\pm$2.70 \\
& & PCC~$\uparrow$  & 0.325$\pm$0.132 & -0.166$\pm$0.138 & 0.152$\pm$0.138 & -0.044$\pm$0.140 & 0.436$\pm$0.130 & 0.248$\pm$0.141 & -0.024$\pm$0.149 & 0.394$\pm$0.129 \\
& & SNR (dB)~$\uparrow$ & -4.57$\pm$0.36 & -6.57$\pm$0.60 & -6.76$\pm$0.59 & -6.13$\pm$0.54 & -4.83$\pm$0.44 & -4.00$\pm$0.47 & -5.38$\pm$0.34 & -4.17$\pm$0.36 \\

\bottomrule
\end{tabular}
\end{adjustbox}
    \caption{Evaluation results. Unsupervised models: Spatial Average, Landmarker, MCSS, Full-body, and SkinMap. Supervised models: DeepPhys, EfficientPhys, and PhysFormer. The best MAE are highlighted in bold and (*) indicates the worst values}
    \label{performance_comparison}
\end{table*}

\cref{performance_comparison} gives a detailed comparison of models on the SYNC-rPPG and UBFC-Phys datasets. SkinMap outperforms other models in the head rotation task in SYNC-rPPG and in the talking and arithmetic tasks in UBFC-Phys. It stays competitive in stable scenarios and maintains its accuracy during challenging tasks, showing robustness. Pre-trained supervised models, especially EfficientPhys, excel in the rest scenario, where the samples closely match the UBFC-rPPG dataset on which they were trained. However, they fail in the talking scenario, with MAE nearly twice that of SkinMap. SkinMap's superior performance comes from its ability to adjust weights during a video to use neck regions when the face is partly hidden. In contrast, Spatial Averaging, Landmarker, and MCSS perform well in stationary tasks but struggle in challenging real-world scenarios, particularly Landmarker. The full-body model, a non-weighted version of SkinMap, performs robustly but still under-performs compared to SkinMap, highlighting the value of the weighting mechanism. This evaluation suggests that while simple ROIs selection or supervised approaches may suffice for static conditions, they are insufficient for real-life applications. For practical use, models must leverage all available sources of information. SkinMap, which does not require additional face detection or extensive pre-processing, offers a robust solution by prioritizing skin pixels, therefore filtering out sudden environmental noise from body movements and lighting variations, thereby enhancing signal quality.

\section{Conclusions and Future Works}
\label{sec:conclusions}

This study presents SkinMap, a full-body skin segmentation model that utilizes all available skin regions and assigns optimized pixel-wise weights for unsupervised rPPG signal extraction pipelines. A new video-PPG dataset was collected at a uniform sampling rate across four real-world scenarios. SkinMap was trained using a synthesized dataset of image–mask pairs. Experimental results indicate that SkinMap accurately detects skin regions and generalizes effectively across diverse skin tones, while distinguishing non-skin areas such as accessories and hair. Compared with existing skin segmentation, ROIs selection approaches, and state-of-the-art supervised methods, SkinMap demonstrates superior performance and robustness in complex, dynamic scenarios. Future work will focus on reducing the model size for deployment on mobile devices.

{
    \small
    \bibliographystyle{ieeenat_fullname}
    \bibliography{main}
}

% WARNING: do not forget to delete the supplementary pages from your submission 
% \input{sec/x_supp}

\end{document}